\documentclass[conference]{IEEEtran}
\IEEEoverridecommandlockouts

\usepackage{cite}
\usepackage{amsmath,amssymb,amsfonts}
\usepackage{algorithmic}
\usepackage{graphicx}
\usepackage{textcomp}
\usepackage{xcolor}

\usepackage{color}
\usepackage[bookmarks=true,hidelinks]{hyperref}
\usepackage{siunitx}
\usepackage{subfig}
\usepackage{booktabs}
\usepackage{ifthen} 

\def\BibTeX{{\rm B\kern-.05em{\sc i\kern-.025em b}\kern-.08em
    T\kern-.1667em\lower.7ex\hbox{E}\kern-.125emX}}

\newcommand{\probP}{\text{I\kern-0.15em P}}

\newcommand{\Cpp}{C\raise.08ex\hbox{\tt ++}}

\newcommand{\ignore}[1]{}

\newboolean{HIDENOTES}
\setboolean{HIDENOTES}{false}

\ifthenelse{\boolean{HIDENOTES}}{
    \newcommand{\OS}[1]{{}}
    \newcommand{\DZ}[1]{{}}
    \newcommand{\AB}[1]{{}}
    \newcommand{\AW}[1]{{}}
    \newcommand{\CONT}[1]{{}}
}{
    \newcommand{\OS}[1]{\textcolor{red}{#1}}
    \newcommand{\DZ}[1]{\textcolor{blue}{#1}}    
    \newcommand{\AB}[1]{{\textcolor{green}{#1}}}    
    \newcommand{\AW}[1]{{\textcolor{yellow}{#1}}}

}

\newcommand{\ARXIVappendixref}{the appendix~(Sec.~\ref{sec:app})}
\newcommand{\ICRAappendixref}{the extended version of this paper~\cite{DBLP:journals/corr/abs-2503-17846}}
\newboolean{isARXIVversion}

\begin{document}

\setboolean{isARXIVversion}{true}
\newcommand{\RefAppendix}{\ifthenelse{\boolean{isARXIVversion}}{\ARXIVappendixref}{\ICRAappendixref}}

\title{Smart Ankleband for Plug-and-Play Hand-Prosthetic Control}

\author{\IEEEauthorblockN{Dean Zadok$^{1}$, Oren Salzman$^{1}$, Alon Wolf$^{2}$, Alex M. Bronstein$^{1}$}
\IEEEauthorblockA{\textit{$^{1}$Dept. of Computer Science, $^{2}$Dept. of Mechanical Engineering} \\
\textit{Technion - IIT}\\
Haifa, Israel \\
\{deanzadok,osalzman,bron\}@cs.technion.ac.il, alonw@me.technion.ac.il}
}

\maketitle

\begin{abstract}
Building robotic prostheses requires  a sensor-based interface designed to provide the robotic hand with the control required to perform hand gestures.
Traditional Electromyography (EMG) based prosthetics and emerging alternatives often face limitations such as muscle-activation limitations, high cost, and complex calibrations.
In this paper, we present a low-cost robotic system composed of a smart ankleband for intuitive, calibration-free control of a robotic hand, and a robotic prosthetic hand that executes actions corresponding to leg gestures.
The ankleband integrates an Inertial Measurement Unit (IMU) sensor with a lightweight neural network to infer user-intended leg gestures from motion data.
Our system represents a significant step towards higher adoption rates of robotic prostheses among arm amputees, as it enables one to operate a prosthetic hand using a low-cost, low-power, and calibration-free solution.
To evaluate our work, we collected data from 10 subjects and tested our prototype ankleband with a robotic hand on an individual with an upper-limb amputation. 
Our results demonstrate that this system empowers users to perform daily tasks more efficiently, requiring few compensatory movements.
\end{abstract}



\section{Introduction}
\label{sec:introduction}

People with upper-limb differences, especially amputees, struggle with Activities of Daily Living (ADLs)~\cite{edemekong2019activities} due to the limitations of current prosthetics, which often lack functionality, intuitive control, and can cause discomfort or pain. 
Despite the dominance of muscle-based control technologies like electromyography (EMG)~\cite{li2021gesture, simpetru2023proportional} and ultrasound (US)~\cite{DBLP:conf/chi/McIntoshMFP17,ZadokSWB23}, their limitations, including muscle disuse, pain, and the need for complex training, result in low adoption rates of current prosthetics, decreased quality of life, and reluctance to experiment with newer developments.

In recent years, researchers have begun to explore alternative control methods for prosthetic arms that do not involve muscle activation.
%
These include autonomous behaviors such as predicting the action type using motion~\cite{uyanik2019deep, DBLP:journals/sensors/CuiLDYL22}, detecting objects through vision sensing~\cite{DBLP:journals/finr/CastroD22, DBLP:journals/corr/abs-2407-12807}, providing instructions using voice commands~\cite{asyali2011design}, and even control robotic devices through the interpretation of brain activity via electroencephalography (EEG)~\cite{maibam2024enhancing}.
However, such methods face various operational limitations. For example, environmental factors such as lighting and motion may degrade the precision of vision-based methods, and to process high-resolution data, one will need expensive wearable hardware to allow precise prosthetic control.
Addressing these limitations, including the discomfort associated with muscle-based control, is crucial for increasing the adoption of robotic prosthetic arms.

\begin{figure}[t]
\centering
\includegraphics[trim=0 0 0 0, clip, width = 0.425\textwidth]{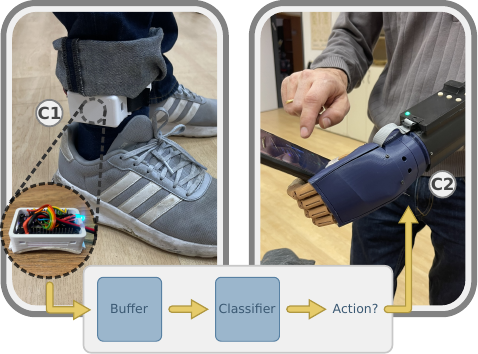}
\vspace{-0.1cm}
\caption{
Our robotic system is comprised of two components:
\textbf{C1}~smart ankleband (left)
and
\textbf{C2}~robotic hand (right).
An IMU placed in the smart ankleband collects and stores signals in a buffer which serves as an input to a machine-learning model able to classify user leg gestures.
The corresponding action is transmitted to the robotic hand.}
\label{fig:main_diagram}
\end{figure}

In response to these challenges, this paper presents an end-to-end low budget, low power, wearable system that allow upper-extremity amputees to seamlessly control a robotic hand  without calibration.\footnote{Our code and dataset are at \textcolor{blue}{\href{https://github.com/deanzadok/ankleband}{https://github.com/deanzadok/ankleband}}.}
Our system (Fig.~\ref{fig:main_diagram}) is based on a smart ankleband, composed of a lightweight microcontroller, a battery, and an IMU.
The ankleband uses a pre-trained machine-learning model to infer leg gestures that the user performs, which correspond to predefined actions, given the motion data generated from the IMU. These are then transmitted to the robotic hand using a wireless interface which, in turn, executes the relevant action.
In our empirical evaluation, we demonstrate the ability of our system to accurately infer gestures for unseen individuals, by collecting data from $10$ subjects.
Finally, we demonstrate via a user study in the lab how the system can be used by {a user with an upper-limb amputation} and enhance the performance of ADL.


\section{SYSTEM DESIGN}
\label{sec:sys-design}

\subsection{Overview}

Our design choices aim to increase the adoption of assistive devices for amputees are guided by three primary objectives:

\begin{itemize}
    \item[\textbf{O1}] Affordability---the system cost should be minimized as this is a significant barrier to adoption.
    \item[\textbf{O2}] Reliability---the system should exhibit high performance even under low-energy and cost constraints.
    \item[\textbf{O3}] User independent---the system should be ``plug and play'', eliminating the need for  calibration for new users.
\end{itemize}

To this end, we propose a robotic system (Fig.~\ref{fig:main_diagram}) consisting of 
component \textbf{C1}---a wearable device placed on the ankle called the \emph{smart ankleband}
and
component \textbf{C2}---a robotic hand.
The smart ankleband  is designed to
detect intended leg gestures performed by the user
and
transmit them to the robotic hand using a wireless interface which, in turn, executes the corresponding action.
The smart ankleband (Fig.~\ref{fig:main_diagram}, left) includes an IMU motion sensor for motion detection, a microcontroller for leg-gesture classification, Bluetooth Low-Energy (BLE) command transmission, and a power source.
The robotic hand (Fig.~\ref{fig:main_diagram}, right) is designed to be generic and can support a wide range of platforms. We based our choice on an open-source project designed to make robotic hands accessible and programmable\footnote{Full details on how to rebuild and program the hand are in \textcolor{blue}{\href{https://github.com/Haifa3D/hand-mechanical-design}{https://github.com/Haifa3D/hand-mechanical-design}}.}.

\begin{figure}[t]
\centering
\includegraphics[trim=0 0 0 0, clip, width = 0.425\textwidth]{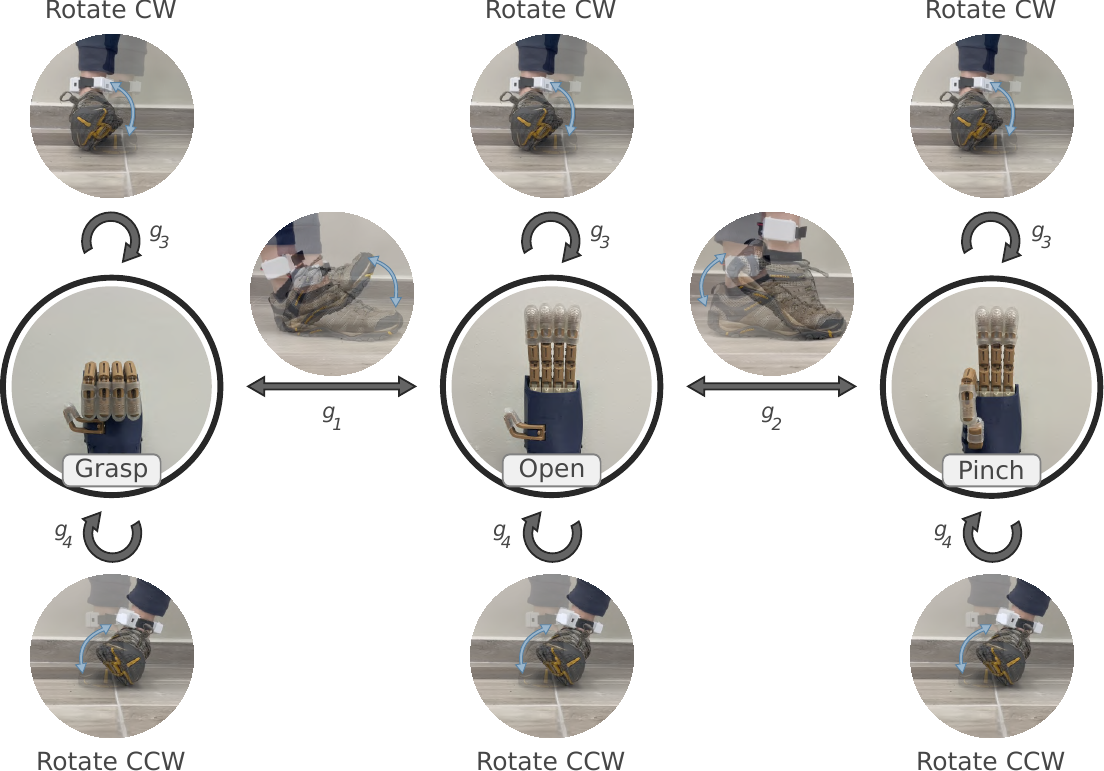}
\vspace{-0.1cm}
\caption{
State machine depicting leg gestures and corresponding hand actions.
We start with an open hand, and the user can either grasp (leg gesture $g_1$), or pinch (leg gesture $g_2$).
To rotate the hand clockwise (CW) or counterclockwise (CCW), the user can use leg gestures~$g_3$ or~$g_4$, respectively.
Rotation is available regardless of the state.}
\label{fig:gestures_demonstration}
\vspace{-0.3cm}
\end{figure}

Based on user feedback, we define four robotic hand actions: grasp, point, rotate wrist left, and rotate wrist right (Fig.~\ref{fig:gestures_demonstration}).
Accordingly, we define four different leg gestures corresponding to each action. 
To return to an open-hand state, the user should perform the last-performed gesture.
Importantly, the specific leg gestures were selected based on the spatial displacement of the IMU sensor, with the four distinct gestures designed to simplify the classification process (Fig.~\ref{fig:gestures_demonstration}). While the gestures are similar to those used in previous papers adopting leg-gesture classification~\cite{zhu2022recognizing}, we require each gesture to be performed twice to minimize unintended activations.

\subsection{Smart-Ankleband Physical Design}
\label{sec:smart-ankleband-physical-design}

The smart ankleband integrates a microcontroller, IMU, and power source into a comfortable and compact wearable design (Fig.~\ref{fig:main_diagram}, left) that requires no additional clothing or user-specific adaptations.
To optimize comfort and weight distribution, the microcontroller and IMU are housed in one component, with the power source in a separate one.

To achieve low cost ($\textbf{O1}$) while ensuring reliability ($\textbf{O2}$), we selected the ESP32 microcontroller for its balance of processing power, varied interfaces, and energy efficiency, and the Adafruit BNO08X IMU for its cost-effectiveness and ability to generate high-frequency 3D acceleration and angular velocity data (up to~\SI{1}{\kilo\hertz}).
Finally, the power source is a 3.7V battery with~950 mAh. 
The overall cost of the smart ankleband is roughly~30~USD, requiring only standard 3D printing and basic soldering.


\subsection{Leg-Gesture Classification}
\label{ssec:leg-gesture-classifier}

Recall that we predefine a set of leg gestures to control the robotic hand.
%
Specifically, at time $t$ the IMU outputs a vector $x_t = [ \ddot{x}_t, \ddot{y}_t, \ddot{z}_t , \dot{\theta}_t^r, \dot{\theta}_t^p, \dot{\theta}_t^y ]$.
Here, $\ddot{x}_t, \ddot{y}_t, \ddot{z}_t$ are the accelerations in the $x$, $y$, and $z$ axes, and $\dot{\theta}_t^r, \dot{\theta}_t^p, \dot{\theta}_t^y$ are the roll, pitch, and yaw angular velocities.
The input~$\bar{x}_t$ is a sequence of $k$ consecutive IMU signals recorded between times $t$ and $t+k-1$.\footnote{The value $k$ is a hyper parameter discussed in Sec.~\ref{sec:evaluation}. It should be large enough to capture leg gestures which typically take between $0.5$ to $1$ seconds but small enough to be processed in real time and include at most one gesture.} Namely 
$\bar{x}_t := \langle x_{t}, \ldots, x_{t+k-1} \rangle$.
The output $p_t$ is a vector of gesture probabilities at time $t$. Specifically, 
$p_t := \langle p^0_t, \ldots, p^4_t \rangle$, 
where 
$p^0_t$ is the probability for no gesture
and
$p^i_t$ for $i\in \{1, \ldots 4 \}$ represents the probability of gesture $g_i$ being performed at time $t$.

As the user can perform a leg gesture at any given time and its length varies, we need to carefully associate the input~$\bar{x}_t$ to our classifier and the continuous time interval~$[t_s^g, t_e^g ]$ duration during which leg gesture $g$ was performed.
With a slight abuse of notation, given the time interval~$[t_s^g, t_e^g ]$ for leg gesture $g$ and the input sequence $\bar{x}_t$, we define $g(\bar{x}_t,t_s^g, t_e^g)$ to be the proportion of time where gesture $g$ lies in the interval.
Namely, 
\begin{equation}
    \label{eq:sigma}
    g(\bar{x}_t,t_s^g, t_e^g) := \frac{[t_s^g, t_e^g] \cap [t, t+k-1]}{ [t_s^g, t_e^g] \cup [t,t+k-1]}.
\end{equation}
Now, given a threshold $\sigma$ (chosen empirically as we detail in Sec.~\ref{sec:model-study}), we label gesture $g$ in interval~$[t_s^g, t_e^g ]$ if $g(\bar{x}_t,t_s^g, t_e^g) \geq \sigma$.
Notice that $g(\bar{x}_t,t_s^g, t_e^g) = 1$ if $g$ is fully present within~$\bar{x}_t$ (i.e., $t \leq t_s^g$ and $t_e^g \leq t+k-1$), and 
$g(\bar{x}_t,t_s^g, t_e^g) = 0$ if~$g$ is not present at all (i.e., $t_s^g < t$ or $t+k-1 < t_e^g$).
Our compact neural network includes a 1D convolutional layer for temporal feature extraction, followed by batch normalization to enhance performance for unseen users, and a multi-layer perceptron to generate the gesture probability vector $p_t$. 
%
Additional technical details can be found in \RefAppendix.

\begin{figure*}[t]
\centering
\subfloat[\label{fig:seq_length}]{\includegraphics[trim=0 0 0 0, clip, width=0.22\textwidth]{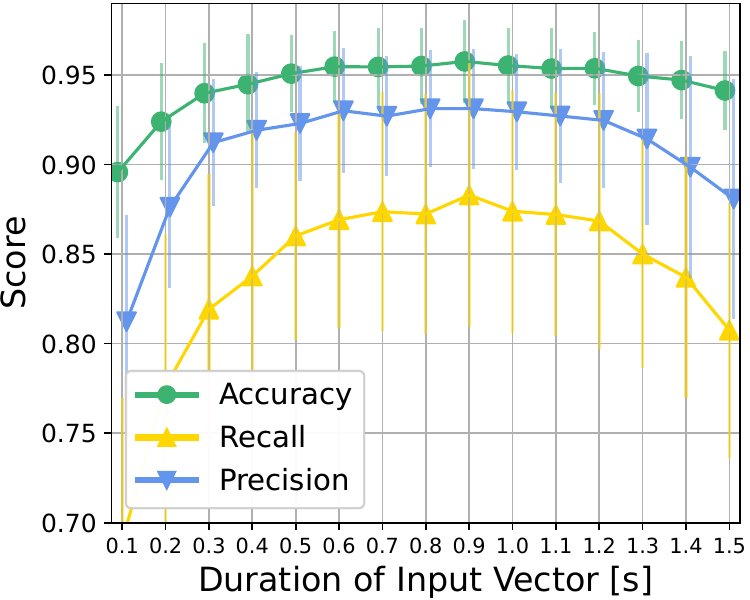}}\hspace{0.2cm}
\subfloat[\label{fig:gesture_percentage}]{\includegraphics[trim=0 0 0 0, clip, width=0.22\textwidth]{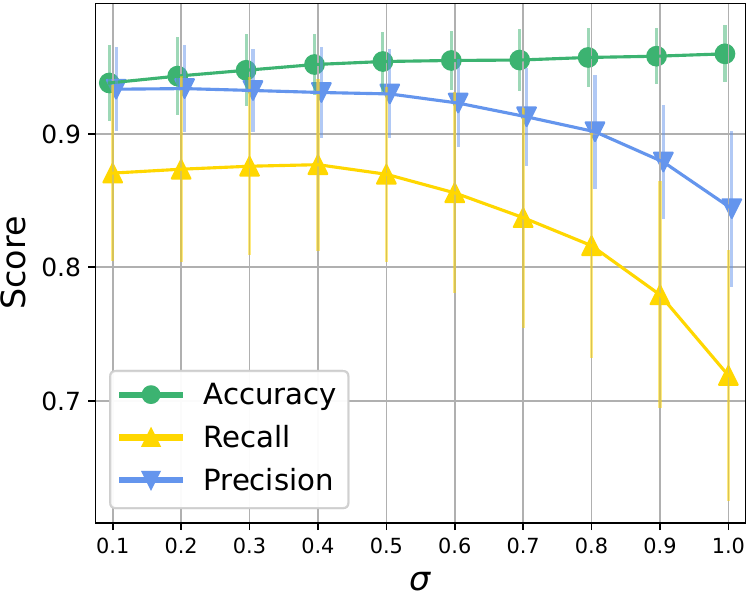}}\hspace{0.2cm}
\subfloat[\label{fig:methods}]{\includegraphics[trim=0 0 0 0, clip, width=0.264\textwidth]{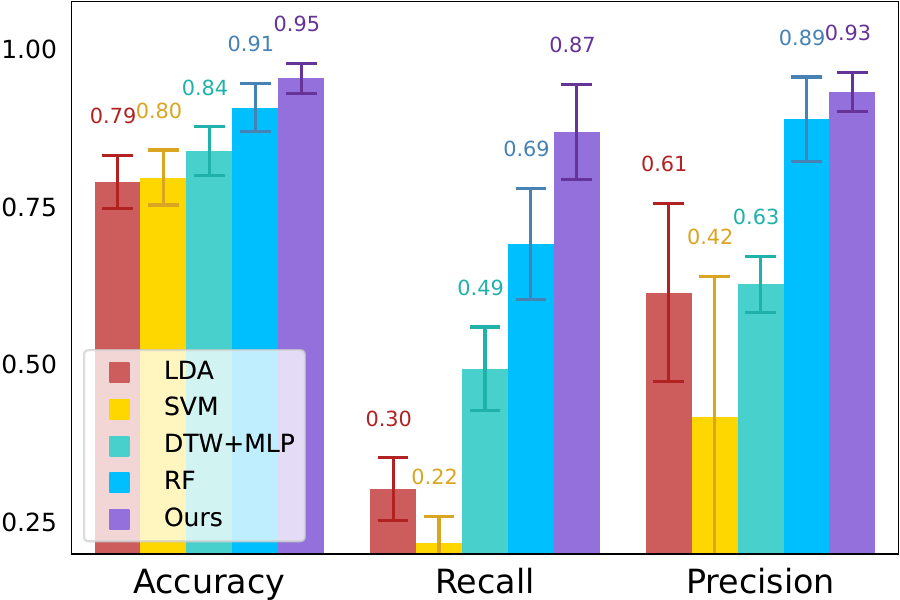}}\hspace{0.2cm}
\subfloat[\label{fig:conf_matrix}]{\includegraphics[trim=10 10 10 10, clip, width=0.24\textwidth]{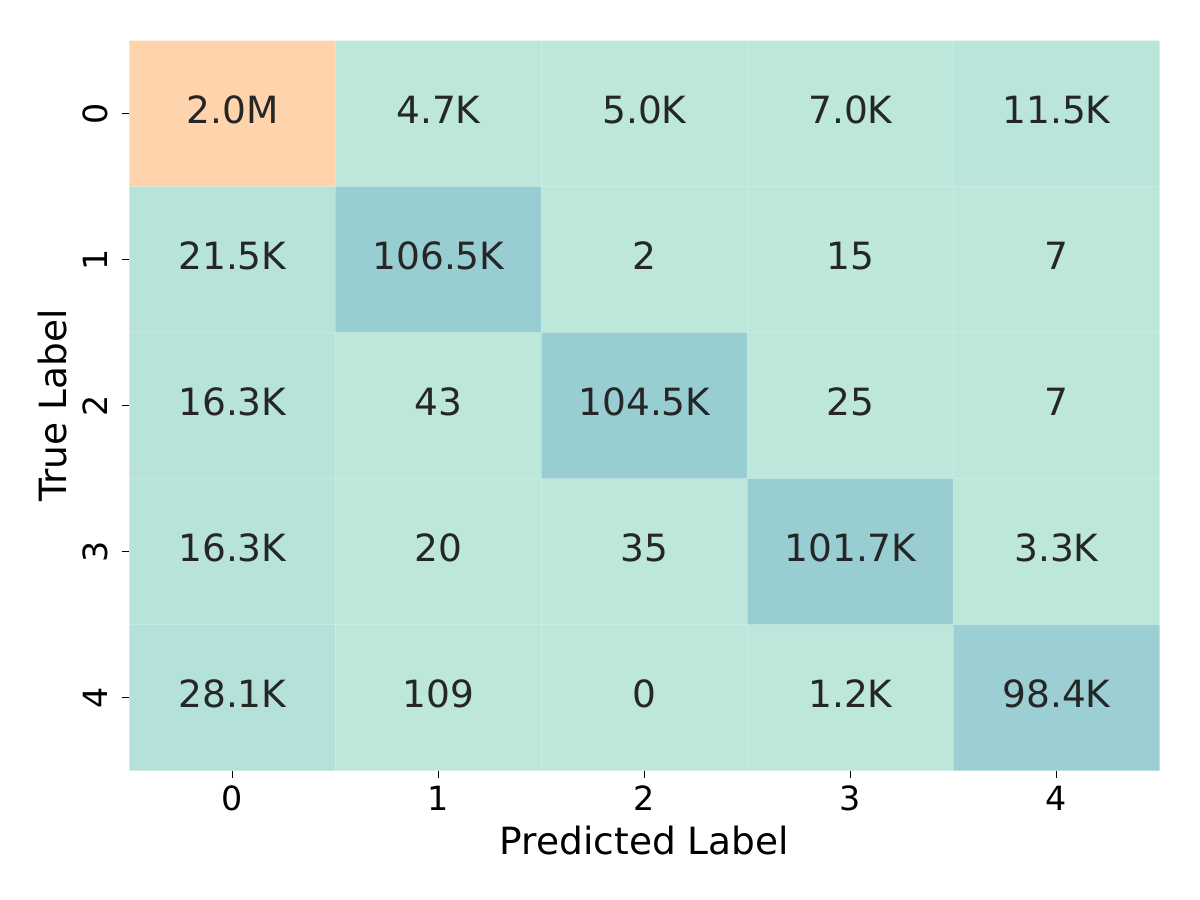}}
\vspace{-0.2cm}
\caption{
(a)~Performance as a function of the duration of the input vector in seconds.
(b)~Performance as a function of $\sigma$ (the minimum percentage of gesture overlap with the input window, Sec.~\ref{ssec:leg-gesture-classifier}). 
(c)~Evaluation of popular classification methods in comparison to the chosen method.
(d)~Confusion matrix for the five classes: 
a no-gesture class
and the 
four gestures. Rows and columns represent true labels and predictions, respectively.
All experiments are averaged over $10$ folds, with one subject left out for test set in each fold. Vertical segments denote one standard deviation.
}
\vspace{-0.5cm}
\end{figure*}

\section{EVALUATION}
\label{sec:evaluation}


We collected data from $10$ subjects (five men, five women, average age 27) free of neurological leg disorders.
All subjects wore the smart ankleband on their right leg, while a Vicon motion-capture system monitored foot motion to compute label intervals after recordings. 
Subjects repeatedly performed each gesture for 1-2 minutes, both while sitting and standing, with each session repeated twice.
In addition, we collected two minutes of ``noise'' data (no specific leg gesture) to improve model robustness. This process resulted in $8$-$16$ labeled minutes of leg gestures and four minutes of regular activity per subject.
The process was approved by the institution's Ethics Committee.
The overall dataset includes a total of $2.5$ million~$x_t$ samples. For each sample, we create the input vector $\bar{x_t}$ and label it to receive the tuple $\langle \bar{x_t}, p_t \rangle$ defined in Sec.~\ref{ssec:leg-gesture-classifier}. 
On feedforward, our model receives a sequence of $k=60$ IMU samples~$\bar{x}_t$, and output the gesture probabilities $p_t$.
For technical details regarding model training and architecture, please refer to \RefAppendix.

\subsection{Model Study}
\label{sec:model-study}

For all experiments, we evaluated our models using $10$-fold cross-validation to simulate testing on a new, unseen user.
In this setup, data from a single subject is reserved for testing, while the remaining data from the nine subjects is used for training. This process is repeated for each subject, and the results are the average of the $10$ training sessions.

\textbf{What is the duration for the input window?}
Our classifier takes a vector~$\bar{x}$ of $k$ IMU signals as input, designed to capture leg gestures lasting $0.5$ to $1$ second.
Thus, we evaluated model performance for different durations of $\bar{x}$ ranging from $0.1$ to~$1.5$ seconds (Fig.~\ref{fig:seq_length}), and found the optimal results to be between $0.6$ and $1.0$ seconds.
Considering memory constraints, which limit the size of the tensors during inference, we set the sensor frequency to \SI{100}{\hertz} and chose a $0.6$-second duration, resulting in $k=60$ IMU samples.

\textbf{How should we determine the existence of labels in input windows?}
We optimized the threshold $\sigma$, determining whether a leg gesture that partially exists in the input window should be labeled as present or not (Eq.~\eqref{eq:sigma}). 
Here, we evaluated $\sigma$ values between~$0.1$ and $1.0$ (Fig.~\ref{fig:gesture_percentage}), and found that higher values significantly decreased performance.
However, to ensure that each input window contains only one prominent gesture, we require $g(\bar{x}_t,t_s^g, t_e^g) \geq 0.5$. Thus, we set $\sigma = 0.5$.

\textbf{How do we choose the best model?}
The modular design of our system allows for classifier substitution.
Our model is constrained by a dynamic memory limit of \SI{90}{\kilo\byte} due to our limited hardware.
Therefore, larger deep-learning (DL) models adopted in previous studies were not feasible~\cite{uyanik2019deep,DBLP:journals/sensors/CuiLDYL22}.
We evaluated classical data-driven approaches such as Linear Discriminant Analysis (LDA)~\cite{DBLP:conf/healthcom/WangM18}, Support Vector Machine (SVM)~\cite{zhu2022recognizing,tchane2023dynamic}, and Random Forest (RF)~\cite{nadaf2023classifying} (Fig.~\ref{fig:methods}).
Additionally, we experimented with Dynamic Time Warping (DTW)~\cite{tchane2023dynamic,iwana2020time} for feature extraction, followed by a classifier to perform gesture recognition. 
Consequently, our proposed DL model outperformed the other methods, despite its limited size.
A possible explanation is that, unlike high-end sensors, our IMU sensor produces noisy data, requiring denoising capabilities which DL methods excel at~\cite{DBLP:journals/ral/BrossardBB20a, DBLP:journals/siamis/EladKV23}.
%
%
We also explored advanced optimization techniques like domain adaptation~\cite{DBLP:conf/iccv/CarlucciPCRB17} and contrastive learning~\cite{DBLP:conf/iclr/AbbaspourazadEM24} to improve generalization to unseen subjects. However, supervised learning proved most effective in our setting.

We generated the confusion matrix of the entire dataset to see how the model classifies each gesture compared to the ground truth (Fig.~\ref{fig:conf_matrix}).
While gesture $g_4$ presented the most challenge, the model effectively predicted all four gestures.
Crucially, the model avoided misclassifying gestures as other gestures, which ensures safety for tasks requiring reliable recognition.
%
%
For a comprehensive evaluation of our model, please refer to \RefAppendix.


\begin{figure}[t]
\centering
\includegraphics[trim=0 0 0 0, clip, width = 0.468\textwidth]{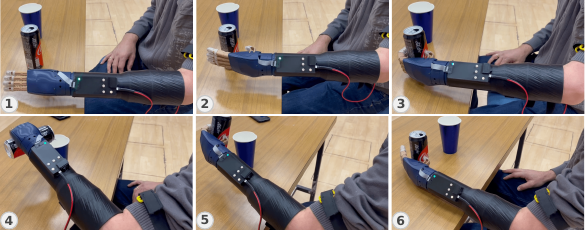} \\ [-0.05ex]
\subfloat{\includegraphics[trim=441 200 350 100, clip, width=0.115\textwidth]{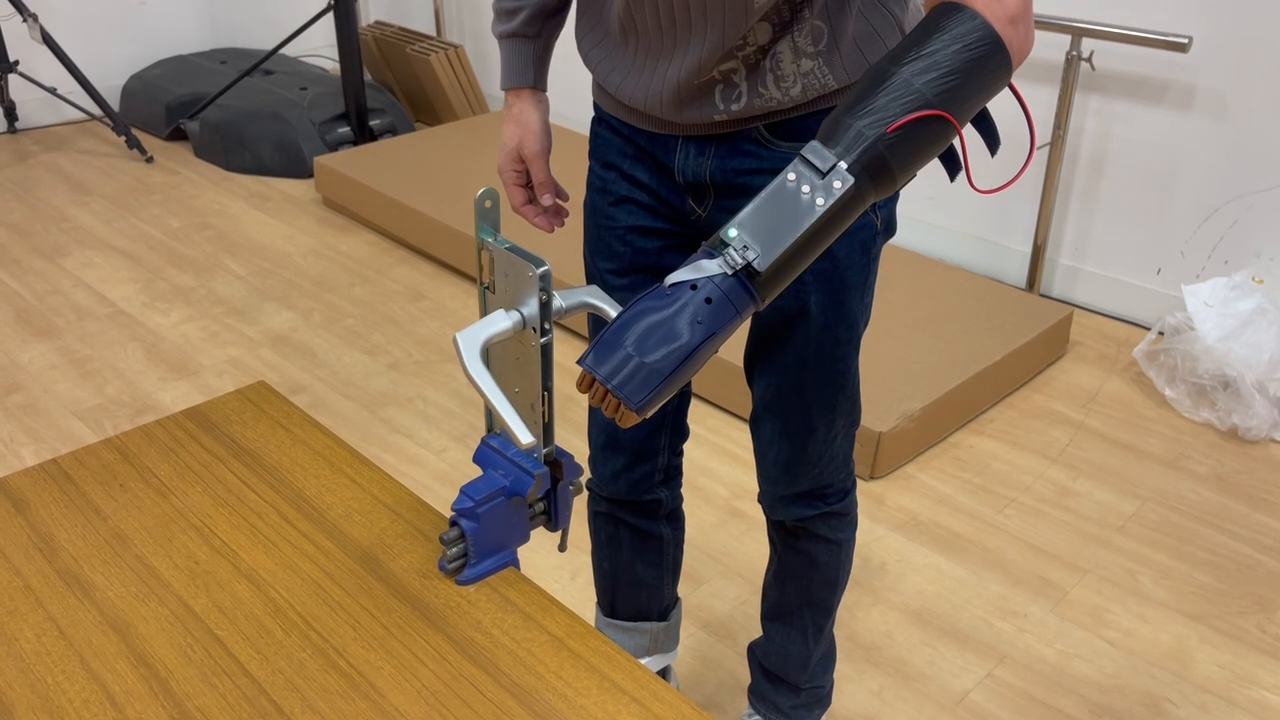}}\hspace{0.5mm}%
\subfloat{\includegraphics[trim=120 100 117 50, clip, width=0.115\textwidth]{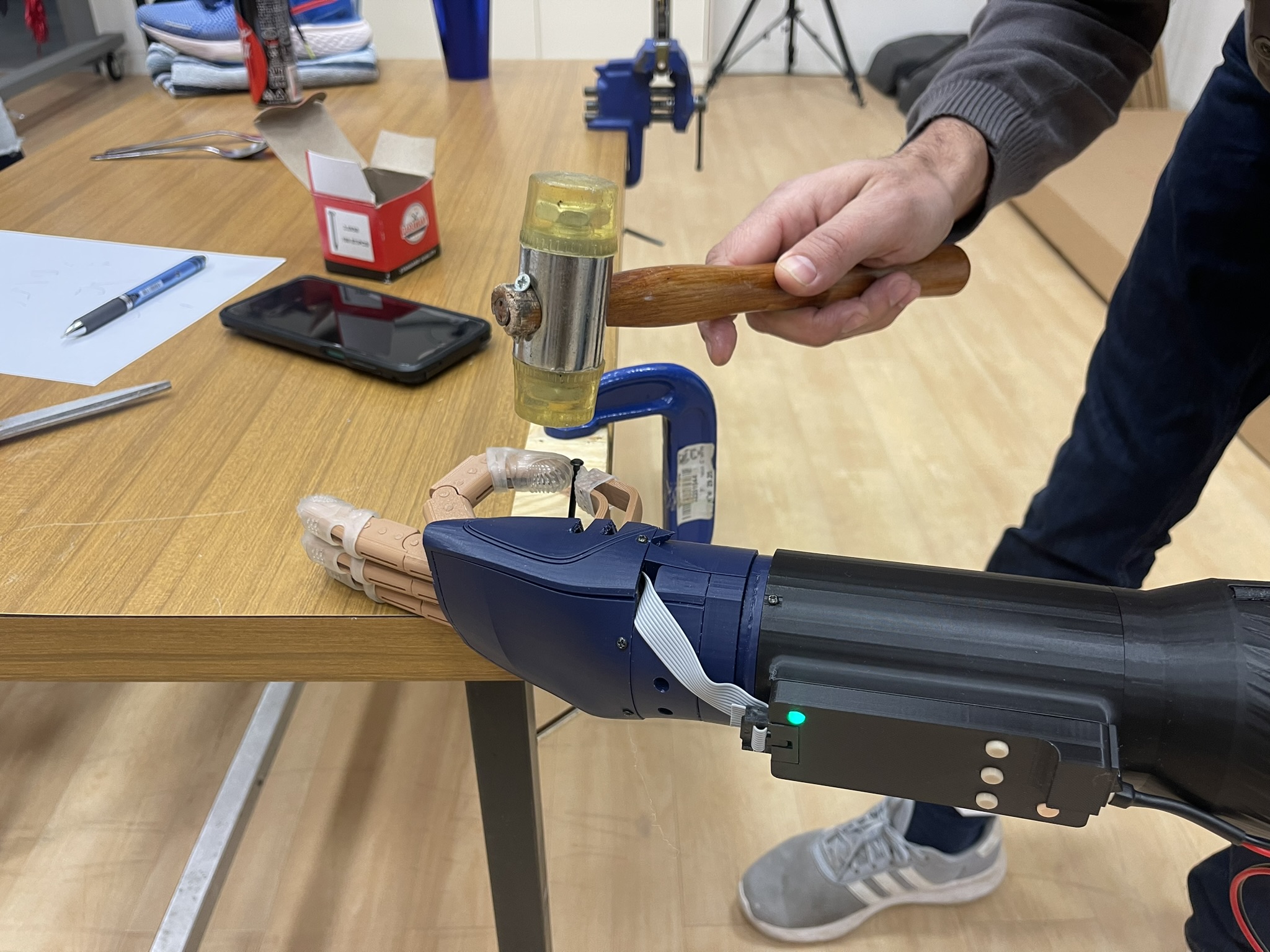}}\hspace{0.5mm}%
\subfloat{\includegraphics[trim=540 158 343 220, clip, width=0.115\textwidth]{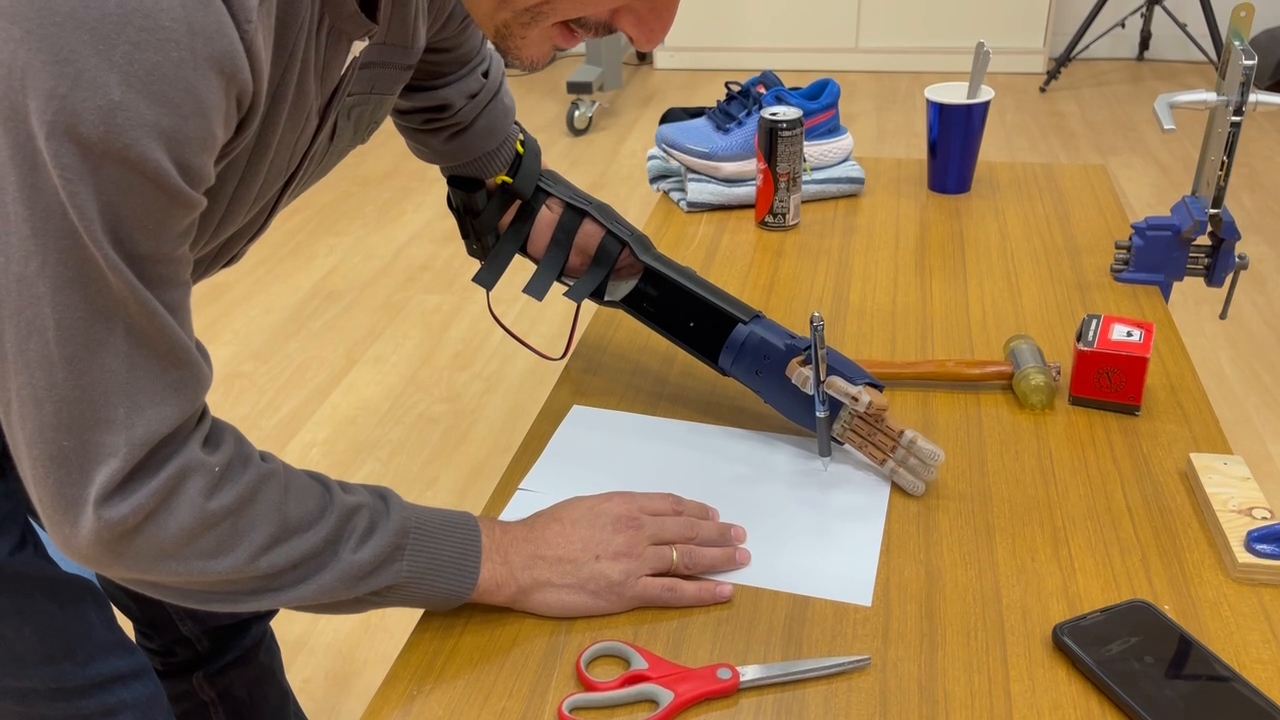}}\hspace{0.5mm}%
\subfloat{\includegraphics[trim=100 160 150 0, clip, width=0.115\textwidth]{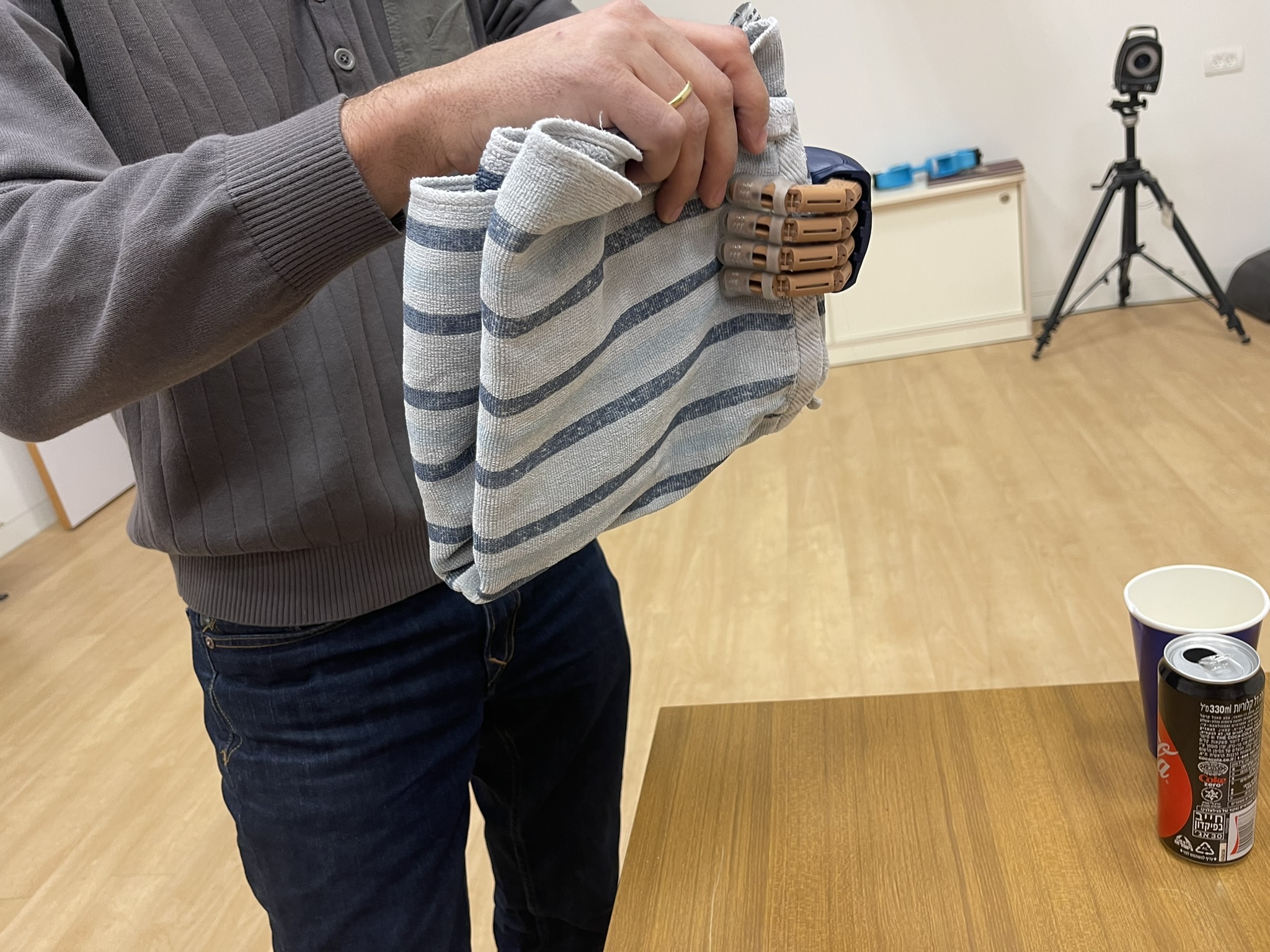}}
\vspace{-0.1cm}
\caption{Examples of tasks completed as part of the AM-ULA test.
(Top) The user reaches the soda can (1) and grasps it using leg gesture $g_1$ (2 and 3). Then, the user lifts and rotates CW using leg gesture $g_3$ to pour the beverage (4). Finally, the user rotates CCW using $g_4$ (5) and opens the hand using $g_1$ to leave the soda can on the table (6).
(Bottom, from left to right) pushing a door knob, hammering a nail, writing on a blank paper, and folding a towel.
}
\label{fig:amula_examples}
\vspace{-0.4cm}
\end{figure}


\subsection{Robotic System Study}
\label{sec:robotic-system-study}

To evaluate our system in ADL, we conducted an in-lab user study with a participant who had left transradial amputation $12$ years ago and had experience with prosthetics. 
The experiment consisted of two tests.
First, the participant performed each leg gesture ten times to familiarize himself with the smart ankleband, without using the robotic hand.
Second, the participant completed the ``Activities Measure for Upper-limb Amputees'' (AM-ULA) test~\cite{resnik2013development}, which consists of $18$ listed activities requiring the operation of a prosthetic hand (Fig.~\ref{fig:amula_examples}).
This test evaluates not only the functionality of our ankleband but also the user's proficiency and interaction with the robotic system.


During the first part, the participant completed $9$ out of $10$ repetitions for the first gesture and $10$ out $10$ for the remaining three gestures, yielding $97.5 \%$ accuracy, despite wearing the ankle band for the first time.
It showcases the intuitive design and minimal training required to use our ankleband.
%
%
For the AM-ULA test, we activated the robotic hand and ensured that the smart ankleband was connected.
Then, we instructed the participant to perform the $18$ tasks, encouraging them to use the robotic hand.
The test revealed that the participant mostly used grasp ($34$ intentions) and pinch ($32$ intentions) actions, with only three instances of wrist rotation. 
The user reported the ankleband to be intuitive and responsive, and that the robotic hand did not perform only a few gestures due to communication malfunctions.
Notably, throughout the two-hour experiment, our ankleband operated reliably and misclassified native movements as gestures only four times, showing promising performance for ADL.


\section{DISCUSSION \& FUTURE WORK}

In this paper, we introduced a ``plug-and-play'' robotic system that uses a smart ankleband to detect intended ankle gestures, enabling seamless control of a robotic prosthetic arm.
Our results demonstrate that this system achieves subject generalization with above $95\%$ accuracy despite hardware constraints, eliminating the need for prolonged personalization.


Future directions for this research include:
(i)~A large-scale study to explore alternative input gestures and application interfaces by leveraging the modularity of our system.
(ii)~Developing a smart ankleband for proportional control of digital or robotic devices, especially for users unable to perform continuous hand motions.
(iii)~Enabling typing motions through our smart ankleband. We aim to provide users with disabilities the full capability to operate digital devices.



\ifthenelse{\boolean{isARXIVversion}}{\section*{Acknowledgment}

This research was supported in part by the Technion Autonomous Systems Program (TASP), the Wynn Family Foundation, the David Himelberg Foundation, the Israeli Ministry of Science \& Technology grants No. 3-17385 and in part by the United States-Israel Binational Science Foundation (BSF) grants no. 2019703 and 2021643.
We also thank the subjects who participated in the data collection, and Haifa3D, a non-profit offering 3D-printed solutions, for their consulting and support through the research.}{}

\ifthenelse{\boolean{isARXIVversion}}{

\section*{APPENDIX}
\label{sec:app}


\subsection{Model Training and Inference}
\label{sec:model-training-and-inference}


Before training, accelerations $\ddot{x}_t, \ddot{y}_t, \ddot{z}_t$ and angular velocities $\dot{\theta}_t^r, \dot{\theta}_t^p, \dot{\theta}_t^y$ were normalized to (approximately) fit the range of $[-1,1]$ by dividing each one using the constants~$c_a=10$ and $c_g=2$, respectively\footnote{The normalization factor for acceleration is higher due to the existence of gravity measurements, which means that at rest, the size of the 3D acceleration vector is approximately $9.81$.}.
On feedforward, our model receives a sequence of $k=60$ IMU samples~$\bar{x}_t$, such that each sample is normalized and fed to the network.
%
%
On backpropagation, for the predicted classes probabilities, the loss function for $N=64$ samples in each batch is the Cross-Entropy Loss, i.e.,
$\mathcal{L}_{CE}~= \sum_{j=1}^{N} \sum_{i=1}^{5} \left[ - g_i \log(p_{j, g_i}) \right]$
where $p_{j,g_i}$ is the predicted probability that sample $j$ belongs to leg gesture~$g_i$.
All training sessions were executed for $10$ epochs, to avoid overfitting. Adam~\cite{DBLP:journals/corr/KingmaB14} was used as the optimization method, with a learning rate of $0.0001$.

For real-time inference, the model was implemented using Eigen in \Cpp, and model weights were extracted from the training procedure and uploaded as constant data arrays to the ESP32 microcontroller. The frequency of model execution during runtime on the ESP32 is approximately \SI{75}{\hertz}.


\subsection*{Model and Baselines Implementation}
\label{ssec:model_and_baselines_implementation}

Our hardware imposes a strict memory constraint of 90 \SI{90}{\kilo\byte} for model deployment, significantly narrowing our selection of suitable classifiers. To operate within this limitation, all alternative methods were tuned to comply with this memory requirement.
The RF implementation, for instance, was specifically tailored by restricting both the tree depth and the number of trees to eight.
Similarly, the SVM was deployed in its linear version.
For Dynamic Time Warping (DTW), we utilized parameters that yielded the highest scores, including $20$ candidates per class.
Overall, our chosen model not only outperformed all alternatives but also achieved this while still fitting within the small memory. Our network architecture is presented in Table~\ref{tab:model_architecture}.


\subsection*{Sensor Sampling frequency}
\label{ssec:sensor_sampling_frequency}



\begin{figure*}[t]
\centering
\subfloat[\label{fig:frequency}]{\includegraphics[trim=0 0 0 0, clip, width=0.24\textwidth]{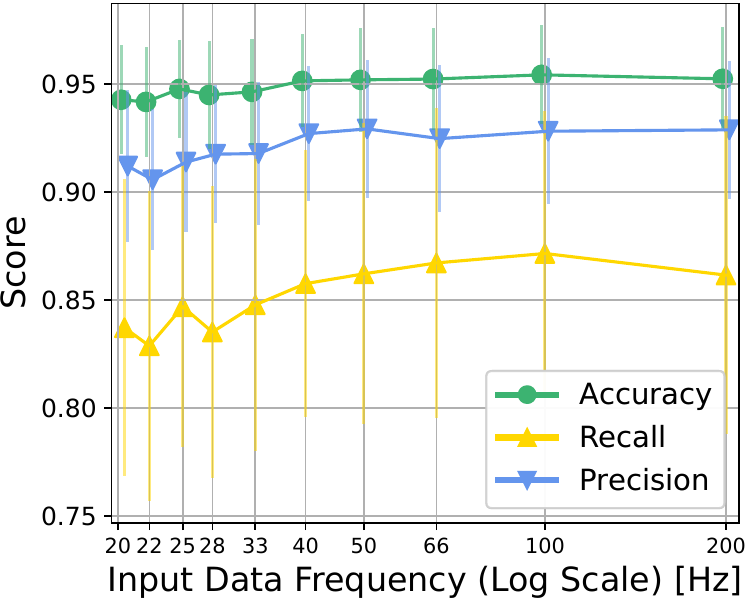}}
\subfloat[\label{fig:num_subjects_train}]{\includegraphics[trim=0 0 0 0, clip, width=0.24\textwidth]{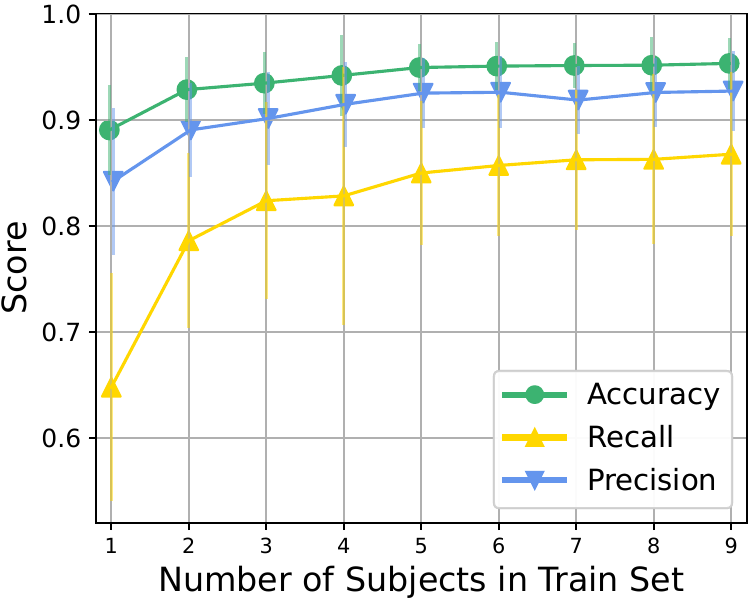}}\hspace{0.2cm}
\subfloat[\label{fig:subjects}]{\includegraphics[trim=0 5 5 5, clip, width=0.485\textwidth]{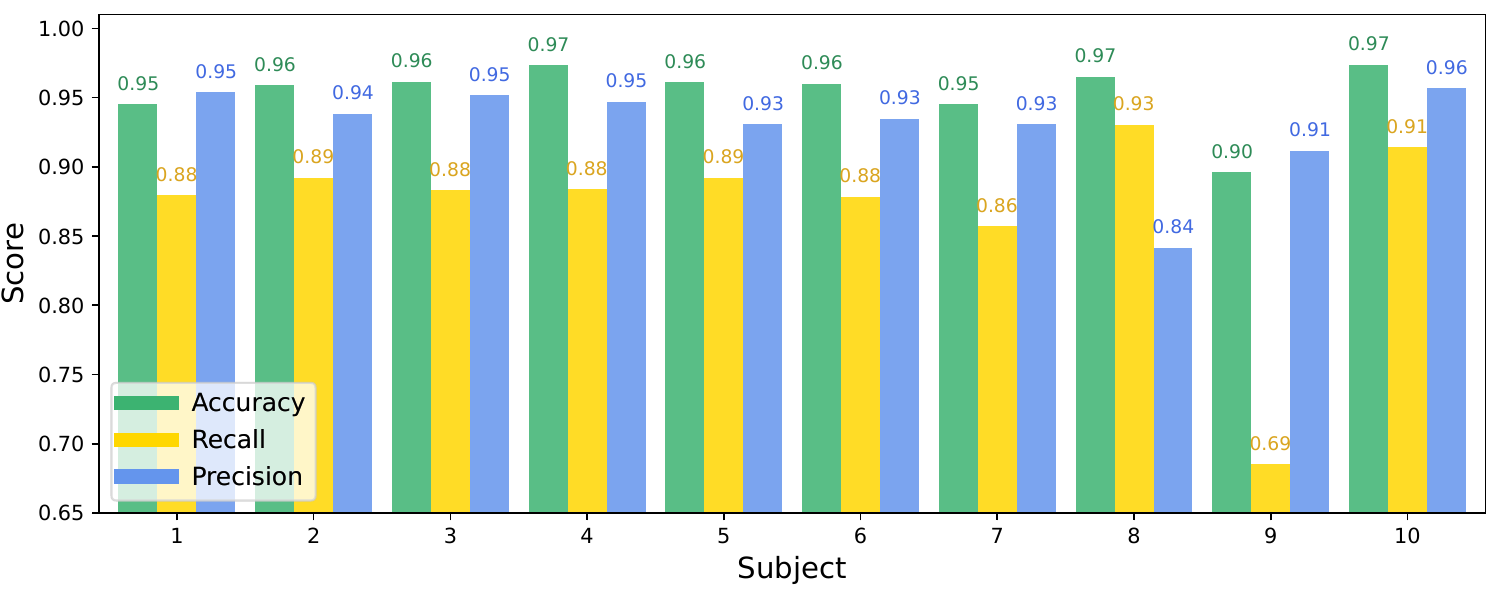}}
\vspace{-0.2cm}
\caption{
(a)~Performance as a function of the sampling frequency of the IMU sensor.
(b)~Performance as a function of the number of subjects in the training set.
(c)~Performance metrics for all ten subjects, plotted using leave-one-subject-out.
Experiments (a) and (b) are averaged over $10$ folds, with one subject left out for the test set in each fold, and a vertical segment denotes one standard deviation.
}
\vspace{-0.4cm}
\end{figure*}

%
The frequency of the sensor used by our model during inference induces a trade-off between reducing the data resolution for faster inference and memory efficiency, and preserving the quality of the input data for potentially better performance. 
Moreover, high-quality performance under a low frequency would allow to choose cheaper hardware (see Sec.~\ref{sec:smart-ankleband-physical-design}).
To address this, we plotted the performance of our model for input frequencies ranging from \SI{20}{\hertz} to \SI{200}{\hertz}.
Results, summarized in Fig.~\ref{fig:frequency}, show that the model improves gradually as the frequency increases.
According to the results and our hardware constraints, we set the sampling frequency to \SI{100}{\hertz}.


\subsection{Subjects Study}
\label{ssec:subjects_study}


Recall that one of the primary objectives of the system design is that it should be ``plug-and-play'' (\textbf{O3}, Sec.~\ref{sec:sys-design}).
To this end, we evaluated the number of subjects required for the model to understand the gestures of a new user.
%
%
For each evaluation, we trained the model, leaving out one subject for a test set and randomly selecting subjects for training from the remaining nine subjects.
The results (Fig.~\ref{fig:num_subjects_train}) indicate that we require only one subject in the training process to detect almost $70\%$ of the gestures performed by the unseen subject (see recall values).
More importantly, the performance improves significantly with the addition of the second subject and continues to improve gradually with the addition of more subjects.
These findings suggest that we can indeed achieve sufficient performance with a relatively small dataset, which makes it practical for real-world deployment.


\begin{table}[t]
\centering
\begin{tabular}{lccc}  
\toprule
Layer & Act. & Out. shape & Parameters \\
\midrule
Conv1D & - & $10 \times 20$ & 180 \\ 
Flatten  & - & $200$ & - \\
Batch Normalization & ReLU & $200$ & 800 \\ 
Linear & - & $64$ & 12,864 \\ 
Batch Normalization & ReLU & $64$ & 256 \\ 
Linear & - & $5$ & 325 \\
\midrule
Total & & & 14,425 \\
\toprule \\ [-4.0ex]
\end{tabular}
\caption{Neural-network architecture of our model.}
\label{tab:model_architecture}
\vspace{-0.4cm}
\end{table}

In addition, we plotted the different metrics for each subject individually to see how efficient the model is across different subjects (Fig.~\ref{fig:subjects}). 
The results show that the model achieved exceptional performance for the majority of the subjects.
However, we observed that when testing the model for subject \#9, the model had difficulties in detecting gesture $g_4$. We associate this gap in performance to the smaller magnitude of motion performed by this subject for this specific gesture, who had limited ability to flex the ankle.
Overall, the method works for all subjects, highlighting the potential for large-scale adoption of our smart ankleband.
}{}

{\small
\bibliographystyle{IEEEtran}
\bibliography{main}

\begin{thebibliography}{10}
\providecommand{\url}[1]{#1}
\csname url@samestyle\endcsname
\providecommand{\newblock}{\relax}
\providecommand{\bibinfo}[2]{#2}
\providecommand{\BIBentrySTDinterwordspacing}{\spaceskip=0pt\relax}
\providecommand{\BIBentryALTinterwordstretchfactor}{4}
\providecommand{\BIBentryALTinterwordspacing}{\spaceskip=\fontdimen2\font plus
\BIBentryALTinterwordstretchfactor\fontdimen3\font minus \fontdimen4\font\relax}
\providecommand{\BIBforeignlanguage}[2]{{%
\expandafter\ifx\csname l@#1\endcsname\relax
\typeout{** WARNING: IEEEtran.bst: No hyphenation pattern has been}%
\typeout{** loaded for the language `#1'. Using the pattern for}%
\typeout{** the default language instead.}%
\else
\language=\csname l@#1\endcsname
\fi
#2}}
\providecommand{\BIBdecl}{\relax}
\BIBdecl

\bibitem{edemekong2019activities}
P.~F. Edemekong, D.~Bomgaars, S.~Sukumaran, and S.~B. Levy, ``Activities of daily living,'' 2019.

\bibitem{li2021gesture}
W.~Li, P.~Shi, and H.~Yu, ``Gesture recognition using surface electromyography and deep learning for prostheses hand: state-of-the-art, challenges, and future,'' \emph{Frontiers in neuroscience}, vol.~15, p. 621885, 2021.

\bibitem{simpetru2023proportional}
R.~C. S{\^\i}mpetru, M.~M{\"a}rz, and A.~D. Vecchio, ``Proportional and simultaneous real-time control of the full human hand from high-density electromyography,'' \emph{{IEEE} Trans. Neural Syst. Rehabilitation Eng.}, 2023.

\bibitem{DBLP:conf/chi/McIntoshMFP17}
J.~McIntosh, A.~Marzo, M.~Fraser, and C.~Phillips, ``Echoflex: Hand gesture recognition using ultrasound imaging,'' in \emph{CHI}, 2017, pp. 1923--1934.

\bibitem{ZadokSWB23}
D.~Zadok, O.~Salzman, A.~Wolf, and A.~M. Bronstein, ``Towards predicting fine finger motions from ultrasound images via kinematic representation,'' in \emph{{ICRA}}, 2023, pp. 12\,645--12\,651.

\bibitem{uyanik2019deep}
C.~Uyanik, S.~F. Hussaini, E.~Erdemir, E.~Kaplanoglu, and A.~Sekmen, ``A deep learning approach for final grasping state determination from motion trajectory of a prosthetic hand,'' \emph{Procedia Comput. Sci.}, vol. 158, pp. 19--26, 2019.

\bibitem{DBLP:journals/sensors/CuiLDYL22}
J.~Cui, Z.~Li, H.~Du, B.~Yan, and P.~Lu, ``Recognition of upper limb action intention based on {IMU},'' \emph{Sensors}, vol.~22, no.~5, p. 1954, 2022.

\bibitem{DBLP:journals/finr/CastroD22}
M.~N. Castro and S.~Dosen, ``Continuous semi-autonomous prosthesis control using a depth sensor on the hand,'' \emph{Front. neurorobot.}, vol.~16, p. 814973, 2022.

\bibitem{DBLP:journals/corr/abs-2407-12807}
M.~A.~B. Sarker, J.~P.~S. Sola, A.~Jones, E.~Laing, E.~Sola{-}Thomas, and M.~H. Imtiaz, ``Vision controlled sensorized prosthetic hand,'' \emph{arXiv}, vol. abs/2407.12807, 2024.

\bibitem{asyali2011design}
M.~H. Asyali, M.~Yilmaz, M.~Tokmakci, K.~Sedef, B.~H. Aksebzeci, and R.~Mittal, ``Design and implementation of a voice-controlled prosthetic hand,'' \emph{Turk. J. Electr. Eng. Comput. Sci.}, vol.~19, no.~1, pp. 33--46, 2011.

\bibitem{maibam2024enhancing}
P.~C. Maibam, D.~Pei, P.~Olikkal, R.~K. Vinjamuri, and N.~M. Kakoty, ``Enhancing prosthetic hand control: A synergistic multi-channel electroencephalogram,'' \emph{Wearable Technologies}, vol.~5, p. e18, 2024.

\bibitem{zhu2022recognizing}
C.~Zhu, L.~Luo, J.~Mai, and Q.~Wang, ``Recognizing continuous multiple degrees of freedom foot movements with inertial sensors,'' \emph{{IEEE} Trans. Neural Syst. Rehabilitation Eng.}, vol.~30, pp. 431--440, 2022.

\bibitem{DBLP:conf/healthcom/WangM18}
Y.~Wang and H.~Ma, ``Real-time continuous gesture recognition with wireless wearable {IMU} sensors,'' in \emph{Healthcom}, 2018, pp. 1--6.

\bibitem{tchane2023dynamic}
G.~V. Tchane~Djogdom, M.~J.-D. Otis, and R.~Meziane, ``Dynamic time warping--based feature selection method for foot gesture cobot operation mode selection,'' \emph{Int. J. Adv. Manuf. Technol.}, vol. 126, no.~9, pp. 4521--4541, 2023.

\bibitem{nadaf2023classifying}
A.~Nadaf and S.~Pardeshi, ``Classifying sign language gestures using decision trees: A comparison of semg and imu sensor data,'' in \emph{INCET}, 2023, pp. 1--8.

\bibitem{iwana2020time}
B.~K. Iwana and S.~Uchida, ``Time series classification using local distance-based features in multi-modal fusion networks,'' \emph{Pattern Recognit.}, vol.~97, p. 107024, 2020.

\bibitem{DBLP:journals/ral/BrossardBB20a}
M.~Brossard, S.~Bonnabel, and A.~Barrau, ``Denoising {IMU} gyroscopes with deep learning for open-loop attitude estimation,'' \emph{{IEEE} Robotics Autom. Lett.}, vol.~5, no.~3, pp. 4796--4803, 2020.

\bibitem{DBLP:journals/siamis/EladKV23}
M.~Elad, B.~Kawar, and G.~Vaksman, ``Image denoising: The deep learning revolution and beyond - {A} survey paper,'' \emph{{SIAM} J. Imaging Sci.}, vol.~16, no.~3, pp. 1594--1654, 2023.

\bibitem{DBLP:conf/iccv/CarlucciPCRB17}
F.~M. Carlucci, L.~Porzi, B.~Caputo, E.~Ricci, and S.~R. Bul{\`{o}}, ``Autodial: Automatic domain alignment layers,'' in \emph{{IEEE} {ICCV}}, 2017, pp. 5077--5085.

\bibitem{DBLP:conf/iclr/AbbaspourazadEM24}
S.~Abbaspourazad, O.~Elachqar, A.~C. Miller, S.~Emrani, U.~Nallasamy, and I.~Shapiro, ``Large-scale training of foundation models for wearable biosignals,'' in \emph{{ICLR}}, 2024.

\bibitem{resnik2013development}
L.~Resnik, L.~Adams, M.~Borgia, J.~Delikat, R.~Disla, C.~Ebner, and L.~S. Walters, ``Development and evaluation of the activities measure for upper limb amputees,'' \emph{Arch. Phys. Med. Rehabil.}, vol.~94, no.~3, pp. 488--494, 2013.

\bibitem{DBLP:journals/corr/KingmaB14}
D.~P. Kingma and J.~Ba, ``Adam: {A} method for stochastic optimization,'' in \emph{{ICLR}}, 2015.

\end{thebibliography}
}


\end{document}